\documentclass[11pt]{article}
\pdfoutput=1
\usepackage{acl2015}
\usepackage{times}
\usepackage{url}
\usepackage{latexsym}
\usepackage{tikz}
\usepackage{color}
\usepackage{float}
\usepackage{amsmath,amsfonts,amssymb}
\usepackage{algorithm}
\usepackage{algpseudocode}
\usepackage{graphicx,subfigure}
\usepackage{booktabs}
\usepackage{multirow}
\usepackage{mystuff-nlp}
\usepackage[normalem]{ulem}
\usepackage{rst}
\newcommand{\jacob}[1]{}
\newcommand{\possen}[1]{\textcolor[rgb]{0,0.3,0}{\uline{#1}}}
\newcommand{\negsen}[1]{\textcolor[rgb]{0.3,0,0}{\uwave{#1}}}

\title{Better Document-level Sentiment Analysis from RST Discourse Parsing\thanks{\:\:Code is available at \protect\url{https://github.com/parry2403/R2N2}}}

\author{Parminder Bhatia \and Yangfeng Ji \and Jacob Eisenstein\\
  School of Interactive Computing \\
  Georgia Institute of Technology \\
  Atlanta, GA 30308\\
  {\tt parminder.bhatia243@gmail.com, \{jiyfeng,jacobe\}@gatech.edu}}

\date{}

\begin{document}
\maketitle
\begin{abstract}
Discourse structure is the hidden link between surface features and document-level properties, such as sentiment polarity. We show that the discourse analyses produced by Rhetorical Structure Theory (RST) parsers can improve document-level sentiment analysis, via composition of local information up the discourse tree. First, we show that reweighting discourse units according to their position in a dependency representation of the rhetorical structure can yield substantial improvements on lexicon-based sentiment analysis. Next, we present a recursive neural network over the RST structure, which offers significant improvements over classification-based methods.
\end{abstract}

\section{Introduction}
\label{sec:intro}
Sentiment analysis and opinion mining are among the most widely-used applications of language technology, impacting both industry and a variety of other academic disciplines~\cite{feldman2013techniques,liu2012sentiment,pang2008opinion}. Yet sentiment analysis is still dominated by bag-of-words approaches, and attempts to include additional linguistic context typically stop at the sentence level~\cite{socher2013recursive}. Since document-level opinion mining inherently involves multi-sentence texts, it seems that analysis of document-level structure should have a role to play.

A classic example of the potential relevance of discourse to sentiment analysis is shown in Figure~\ref{fig:ex-rst}. In this review of the film \emph{The Last Samurai}, the \possen{positive sentiment} words far outnumber the \negsen{negative sentiment} words. But the discourse structure --- indicated here with Rhetorical Structure Theory (RST; Mann and Thompson, 1988)\nocite{mann1984discourse} --- clearly favors the final sentence, whose polarity is negative. This example is illustrative in more than one way: it was originally identified by \newcite{Voll2007Not}, who found that \emph{manually-annotated} RST parse trees improved lexicon-based sentiment analysis, but that automatically-generated parses from the SPADE parser ~\cite{soricut2003sentence}, which was then state-of-the-art, did not.

\begin{figure}
\hspace{0.5in}
\setlength{\compressionWidth}{60pt}
\dirrel
{Concession}
{\dirrel
  {}
  {\rstsegment{\refr{e1}}}
  {Justify}
  {
      \multirel{Conjunction}
      {\dirrel
        {}
        {\rstsegment{\refr{e2}}}
        {Elaboration}
        {\rstsegment{\refr{e3}}}}
      {\rstsegment{\refr{e4}}}
      {\dirrel{}
        {\rstsegment{\refr{e5}}}
        {Justify}
        {\multirel{Conjunction}
          {\rstsegment{\refr{e6}}}
          {\rstsegment{\refr{e7}}}}}
    }
}
{}
{\rstsegment{\refr{e8}}}
\begin{rhetoricaltext}
\footnotesize
\unit[e1]{It could have been a \possen{great} movie}
\unit[e2]{It does have \possen{beautiful} scenery,}
\unit[e3]{some of the \possen{best} since Lord of the Rings.}
\unit[e4]{The acting is \possen{well} done,}
\unit[e5]{and I really \possen{liked} the son of the leader of the Samurai.}
\unit[e6]{He was a \possen{likable} chap,}
\unit[e7]{and I \negsen{hated} to see him die.}
\unit[e8]{But, other than all that, this movie is \negsen{nothing} more than hidden \negsen{rip-offs}.}
\end{rhetoricaltext}
\caption{Example adapted from \newcite{Voll2007Not}.}
\label{fig:ex-rst}
\end{figure}

Since this time, RST discourse parsing has improved considerably, with the best systems now yielding 5-10\% greater raw accuracy than SPADE, depending on the metric. The time is therefore right to reconsider the effectiveness of RST for document-level sentiment analysis. In this paper, we present two different ways of combining RST discourse parses with sentiment analysis. The methods are both relatively simple, and can be used in combination with an ``off the shelf'' discourse parser.
We consider the following two architectures:

\begin{itemize}
\item Reweighting the contribution of each discourse unit, based on its position in a dependency-like representation of the discourse structure. Such weights can be defined using a simple function, or learned from a small of data.
\item Recursively propagating sentiment up through the RST parse, in an architecture inspired by recursive neural networks~\cite{smolensky1990tensor,socher2011parsing}.
\end{itemize}

Both architectures can be used in combination with either a lexicon-based sentiment analyzer, or a trained classifier. Indeed, for users whose starting point is a lexicon-based approach, a simple RST-based reweighting function can offer significant improvements. For those who are willing to train a sentiment classifier, the recursive model yields further gains.

\section{Background}
\label{sec:background}

\subsection{Rhetorical Structure Theory}
RST is a compositional model of discourse structure, in which elementary discourse units (EDUs) are combined intro progressively larger discourse units, ultimately covering the entire document. Discourse relations may involve a \emph{nucleus} and a \emph{satellite}, or they may be \emph{multinuclear}. In the example in Figure~\ref{fig:ex-rst}, the unit $1C$ is the satellite of a relationship with its nucleus $1B$; together they form a larger discourse unit, which is involved in a multinuclear \annot{conjunction} relation. 

The nuclearity structure of RST trees suggests a natural approach to evaluating the importance of segments of text: satellites tend to be less important, and nucleii tend to be more important~\cite{marcu1999automatic}. This idea has been leveraged extensively in document summarization~\cite{gerani2014abstractive,uzeda2010comprehensive,yoshida2014dependency}, and was the inspiration for \newcite{Voll2007Not}, who examined intra-sentential relations, eliminating all words except those in the top-most nucleus within each sentence. More recent work focuses on reweighting each discourse unit depending on the relations in which it participates~\cite{heerschop2011polarity,hogenboom2015using}. We consider such an approach, and compare it with a compositional method, in which sentiment polarity is propagated up the discourse tree.

\newcite{marcu1997rhetorical} provides the seminal work on automatic RST parsing, but there has been a recent spike of interest in this task, with contemporary approaches employing discriminative learning~\cite{hernault2010hilda}, rich features~\cite{feng2012text}, structured prediction~\cite{joty2015codra}, and representation learning~\cite{ji2014representation,li2014recursive}. With many strong systems to choose from, we employ the publicly-available DPLP parser~\cite{ji2014representation},\footnote{\url{https://github.com/jiyfeng/DPLP}}. To our knowledge, this system currently gives the best F-measure on relation identification, the most difficult subtask of RST parsing. DPLP is a shift-reduce parser~\cite{sagae2009analysis}, and its time complexity is linear in the length of the document.

\subsection{Sentiment analysis} 
There is a huge literature on sentiment analysis~\cite{pang2008opinion,liu2012sentiment}, with particular interest in determining the overall sentiment polarity (positive or negative) of a document. Bag-of-words models are widely used for this task, as they offer accuracy that is often very competitive with more complex approaches. Given labeled data, supervised learning can be applied to obtain sentiment weights for each word. However, the effectiveness of supervised sentiment analysis depends on having training data in the same domain as the target, and this is not always possible. Moreover, in social science applications, the desired labels may not correspond directly to positive or negative sentiment, but may focus on other categories, such as politeness~\cite{danescu2013computational}, narrative frames~\cite{jurafsky2014narrative}, or a multidimensional spectrum of emotions~\cite{kim2012you}. In these cases, labeled documents may not be available, so users often employ a simpler method: counting matches against lists of words associated with each category. Such lists may be built manually from introspection, as in LIWC~\cite{tausczik2010psychological} and the General Inquirer~\cite{stone1966general}. Alternatively, they may be induced by bootstrapping from a seed set of words~\cite{hatzivassiloglou1997predicting,taboada2011lexicon}. While lexicon-based methods may be less accurate than supervised classifiers, they are easier to apply to new domains and problem settings. Our proposed approach can be used in combination with either method for sentiment analysis, and in principle, could be directly applied to other document-level categories, such as politeness.

\subsection{Datasets}
\label{sec:back-data}
We evaluate on two review datasets. In both cases, the goal is to correctly classify the opinion polarity as positive or negative. The first dataset is comprised of 2000 movie reviews, gathered by \newcite{pang2004sentimental}. We perform ten-fold cross-validation on this data. The second dataset is larger, consisting of 50,000 movie reviews, gathered by \newcite{socher2013recursive}, with a predefined 50/50 split into training and test sets. Documents are scored on a 1-10 scale, and we treat scores $\leq 4$ as negative, $\geq 7$ as positive, and ignore scores of 5-6 as neutral --- although in principle nothing prevents extension of our approaches to more than two sentiment classes.
\section{Discourse depth reweighting}
\label{sec:dep}
Our first approach to incorporating discourse information into sentiment analysis is based on quantifying the importance of each unit of text in terms of its discourse depth. To do this, we employ the \emph{dependency-based discourse tree (DEP-DT)} formulation from prior work on summarization~\cite{hirao2013single}. The DEP-DT formalism converts the constituent-like RST tree into a directed graph over elementary discourse units (EDUs), in a process that is a close analogue of the transformation of a headed syntactic constituent parse to a syntactic dependency graph~\cite{kubler2009dependency}.

\begin{figure}
  \centering
\begin{tikzpicture}[->,level distance=2em]
\node (1h) {$1H$}
child {node {$1A$}
  child {node {$1B$}
    child {node {$1C$}}
  }
  child {node {$1D$}}
  child {node {$1E$}
    child {node {$1F$}}
    child {node {$1G$}}    
  }
};
\end{tikzpicture}
  \caption{Dependency-based discourse tree representation of the discourse in Figure~\ref{fig:ex-rst}}
  \label{fig:ex-dep}
\end{figure}
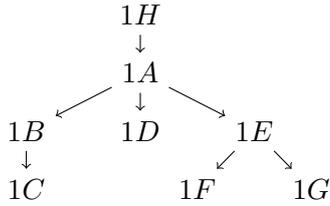

The DEP-DT representation of the discourse in Figure~\ref{fig:ex-rst} in shown in Figure~\ref{fig:ex-dep}. The graph is constructed by propagating ``head'' information up the RST tree; if the elementary discourse unit $e_i$ is the satellite in a discourse relation headed by $e_j$, then there is an edge from $e_j$ to $e_i$. Thus, the ``depth'' of each EDU is the number of times in which it is embedded in the satellite of a discourse relation. The exact algorithm for constructing DEP-DTs is given by \newcite{hirao2013single}.

Given this representation, we construct a simple linear function for weighting the contribution of the EDU at depth $d_i$:
\begin{equation}
\lambda_i = \max(0.5, 1 - d_i / 6).
\end{equation}
Thus, at $d_i = 0$, we have $\lambda_i = 1$, and at $d_i \geq 3$, we have $\lambda_i = 0.5$. Now assume each elementary discourse unit contributes a prediction $\psi_i = \vec{\theta}^{\top}\vec{w}_i$, where $\vec{w}_i$ is the bag-of-words vector, and $\theta$ is a vector of weights, which may be either learned or specified by a sentiment lexicon. Then the overall prediction for a document is given by,
\begin{equation}
\Psi = \sum_i \lambda_i (\vec{\theta}^{\top}\vec{w}_i) =  \vec{\theta}^{\top} (\sum_i \lambda_i \vec{w}_i).
\end{equation}

\paragraph{Evaluation}
We apply this approach in combination with both lexicon-based and classification-based sentiment analysis. We use the lexicon of \newcite{wilson2005recognizing}, and set $\theta_j = 1$ for words marked ``positive'', and $\theta_j = -1$ for words marked negative. For classification-based analysis, we set $\vth$ equal to the weights obtained by training a logistic regression classifier, tuning the regularization coefficient on held-out data.

Results are shown in Table~\ref{tab:results}. As seen in the comparison between lines B1 and D1, discourse depth weighting offers substantial improvements over the bag-of-words approach for lexicon-based sentiment analysis, with raw improvements of $4-5\%$. Given the simplicity of this approach --- which requires only a sentiment lexicon and a discourse parser --- we strongly recommend the application of discourse depth weighting for lexicon-based sentiment analysis at the document level. However, the improvements for the classification-based models are considerably smaller, less than $1\%$ in both datasets.

\jacob{On the P\&L data, Fisher's exact test gives $D1 > B1$, but not $R2 > B2$. We need a sign test for that.}

\begin{table}
  \centering
  \begin{tabular}[tab:results]{lll}
    \toprule
    & Pang \& Lee & Socher et al. \\
    \midrule
    \multicolumn{3}{l}{\emph{Baselines}}\\
    B1. Lexicon & 68.3 & 74.9 \\
    B2. Classifier & 82.4 & 81.5 \\
    \multicolumn{3}{l}{\emph{Discourse depth weighting}}\\
    D1. Lexicon & 72.6 & 78.9 \\
    D2. Classifier & 82.9 & 82.0\\
    \multicolumn{3}{l}{\emph{Rhetorical recursive neural network}}\\
    R1. No relations & 83.4 & 85.5\\
    R2. With relations & 84.1 & 85.6\\
    \bottomrule
  \end{tabular}
  \caption{Sentiment classification accuracies on two movie review datasets~\cite{pang2004sentimental,socher2013recursive}, described in Section~\ref{sec:back-data}.}
  \label{tab:results}
\end{table}

\section{Rhetorical Recursive Neural Networks}
Discourse-depth reweighting offers significant improvements for lexicon-based sentiment analysis, but the improvements over the more accurate classification-based method are meager. We therefore turn to a data-driven approach for combining sentiment analysis with rhetorical structure theory, based on recursive neural networks~\cite{socher2011parsing}. The idea is simple: recursively propagate sentiment scores up the RST tree, until the root of the document is reached. For nucleus-satellite discourse relations, we have:
\begin{equation}
\Psi_i = \tanh(K^{(r_i)}_n \Psi_{n(i)} + K^{(r_i)}_s \Psi_{s(i)}),
\label{eq:psi-ns}
\end{equation}
where $i$ indexes a discourse unit composed from relation $r_i$, $n(i)$ indicates its nucleus, and $s(i)$ indicates its satellite. Returning to the example in Figure~\ref{fig:ex-rst}, the sentiment score for the discourse unit obtained by combining $1B$ and $1C$ is obtained from $\tanh(K^{(\text{elaboration})}_n \Psi_{1B} + K^{(\text{elaboration})}_s \Psi_{1C})$. Similarly, for multinuclear relations, we have, 
\begin{equation}
\Psi_i = \tanh(\sum_{j \in n(i)} K^{(r_i)}_n \Psi_j).
\label{eq:psi-nn}
\end{equation}
In the base case, each elementary discourse unit's sentiment is constructed from its bag-of-words, $\Psi_i = \vtht \vw_i$. Because the structure of each document is different, the network architecture varies in each example; nonetheless, the parameters can be reused across all instances.

This approach, which we call a Rhetorical Recursive Neural Network (R2N2), is reminiscent of the compositional model proposed by \newcite{socher2013recursive}, where composition is over the constituents of the syntactic parse of a sentence, rather than the units of a discourse. However, a crucial difference is that in R2N2s, the elements $\Psi$ and $K$ are \emph{scalars}: we do not attempt to learn a latent distributed representation of the sub-document units. This is because discourse units typically comprise multiple words, so that accurate analysis of the sentiment for elementary discourse units is not so difficult as accurate analysis of individual words. The scores for individual discourse units can be computed from a bag-of-words classifier, or, in future work, from a more complex model such as a recursive or recurrent neural network.

While this neural network structure captures the idea of compositionality over the RST tree, the most deeply embedded discourse units can be heavily down-weighted by the recursive composition (assuming $K_s < K_n$): in the most extreme case of a right-branching or left-branching structure, the recursive operator may be applied $N$ times to the most deeply embedded EDU. In contrast, discourse depth reweighting applies a uniform weight of $0.5$ to all discourse units with depth $\geq 3$. In the spirit of this approach, we add an additional component to the network architecture, capturing the bag-of-words for the entire document. Thus, at the root node we have:
\begin{equation}
\Psi_{\text{doc}} = \gamma \vtht (\sum_i \vw_i) + \Psi_{\text{rst-root}},
\end{equation}
with $\Psi_{\text{rst-root}}$ defined recursively from Equations~\ref{eq:psi-ns} and~\ref{eq:psi-nn}, $\vth$ indicating the vector of per-word weights, and the scalar $\gamma$ controlling the tradeoff between these two components.

\paragraph{Learning} R2N2 is trained by backpropagating from a hinge loss objective; assuming $y_t \in \{-1,1\}$ for each document $t$, we have the loss $L_t = (1 - y_t \Psi_{\text{doc},t})_{+}$. From this loss, we use backpropagation through structure to obtain gradients on the parameters~\cite{goller1996learning}. Training is performed using stochastic gradient descent. For simplicity, we follow \newcite{zirn2011fine} and focus on the distinction between contrastive and non-contrastive relations. The set of contrastive relations includes \annot{Contrast}, \annot{Comparison}, \annot{Antithesis}, \annot{Antithesis-e}, \annot{Consequence-s}, \annot{Concession}, and \annot{Problem-Solution}.

\paragraph{Evaluation} Results for this approach are shown in lines R1 and R2 of Table~\ref{tab:results}. Even without distinguishing between discourse relations, we get an improvement of more than $3\%$ accuracy on the Stanford data, and $0.5\%$ on the smaller Pang \& Lee dataset. Adding sensitivity to discourse relations (distinguishing $K^{(r)}$ for contrastive and non-contrastive relations) offers further improvements on the Pang \& Lee data, outperforming the baseline classifier (D2) by 1.3\%. 

The accuracy of discourse relation detection is only 60\% for even the best systems~\cite{ji2014representation}, which may help to explain why relations do not offer a more substantial boost. An anonymous reviewer recommended evaluating on gold RST parse trees to determine the extent to which improvements in RST parsing might transfer to downstream document analysis. Such an evaluation would seem to require a large corpus of texts with both gold RST parse trees and sentiment polarity labels; the SFU Review Corpus~\cite{taboada2008sfu} of 30 review texts offers a starting point, but is probably too small to train a competitive sentiment analysis system.



\section{Related Work}
Section~\ref{sec:background} mentions some especially relevant prior work. Other efforts to incorporate RST into sentiment analysis have often focused on intra-sentential discourse relations~\cite{heerschop2011polarity,zhou2011unsupervised,chenlo2014rhetorical}, rather than relations over the entire document. \newcite{wang2012exploiting} address sentiment analysis in Chinese. Lacking a discourse parser, they focus on explicit connectives, using a strategy that is related to our discourse depth reweighting. \newcite{wang2013exploiting} use manually-annotated discourse parses in combination with a sentiment lexicon, which is automatically updated based on the discourse structure. \newcite{zirn2011fine} use an RST parser in a Markov Logic Network, with the goal of making polarity predictions at the sub-sentence level, rather than improving document-level prediction. None of the prior work considers the sort of recurrent compositional model presented here.

An alternative to RST is to incorporate ``shallow'' discourse structure, such as the relations from the Penn Discourse Treebank (PDTB). PDTB relations were shown to improve sentence-level sentiment analysis by \newcite{somasundaran2009supervised}, and were incorporated in a model of sentiment flow by \newcite{wachsmuth2014modeling}. PDTB relations are often signaled with explicit discourse connectives, and these may be used as a feature~\cite{trivedi2013discourse,lazaridou2013bayesian} or as posterior constraints~\cite{yang2014context}. This prior work on discourse relations within sentences and between adjacent sentences can be viewed as complementary to our focus on higher-level discourse relations across the entire document.

There are unfortunately few possibilities for direct comparison of our approach against prior work. \newcite{heerschop2011polarity} and \newcite{wachsmuth2014modeling} also employ the \newcite{pang2004sentimental} dataset, but neither of their results are directly comparable: \newcite{heerschop2011polarity} exclude documents that SPADE fails to parse, and \newcite{wachsmuth2014modeling} evaluates only on individual sentences rather than entire documents. The only possible direct comparison is with very recent work from~\newcite{hogenboom2015using}, who employ a weighting scheme that is similar to the approach described in Section~\ref{sec:dep}. They evaluate on the Pang and Lee data, and consider only lexicon-based sentiment analysis, obtaining document-level accuracies between 65\% (for the baseline) and 72\% (for their best discourse-augmented system). Table~\ref{tab:results} shows that fully supervised methods give much stronger performance on this dataset, with accuracies more than 10\% higher.

\section{Conclusion}
\label{sec:con}
Sentiment polarity analysis has typically relied on a ``preponderance of evidence'' strategy, hoping that the words or sentences representing the overall polarity will outweigh those representing counterpoints or rhetorical concessions. However, with the availability of off-the-shelf RST discourse parsers, it is now easy to include document-level structure in sentiment analysis. We show that a simple reweighting approach offers robust advantages in lexicon-based sentiment analysis, and that a recursive neural network can substantially outperform a bag-of-words classifier. Future work will focus on combining models of discourse structure with richer models at the sentence level.
\paragraph{Acknowledgments} 
\begin{small}
Thanks to the anonymous reviewers for their helpful suggestions on how to improve the paper. The following members of the Georgia Tech Computational Linguistics Laboratory offered feedback throughout the research process: Naman Goyal, Vinodh Krishan, Umashanthi Pavalanathan, Ana Smith, Yijie Wang, and Yi Yang. Several class projects in Georgia Tech CS 4650/7650 took alternative approaches to the application of discourse parsing to sentiment analysis, which was informative to this work; thanks particularly to Julia Cochran, Rohit Pathak, Pavan Kumar Ramnath, and Bharadwaj Tanikella. This research was supported by a Google Faculty Research Award, by the National Institutes of Health under award number R01GM112697-01, and by the Air Force Office of Scientific Research. The content is solely the responsibility of the authors and does not necessarily represent the official views of these sponsors.
\end{small}

\bibliographystyle{acl}
\citestr

\end{document}